\documentclass[conference]{IEEEtran}
\IEEEoverridecommandlockouts
\usepackage{cite}
\usepackage{amsmath,amssymb,amsfonts}
\usepackage{algorithmic}
\usepackage{graphicx}
\usepackage{textcomp}
\usepackage{xcolor}
\usepackage{array}
\usepackage{gensymb}
\usepackage{verbatim}
\def\BibTeX{{\rm B\kern-.05em{\sc i\kern-.025em b}\kern-.08em
    T\kern-.1667em\lower.7ex\hbox{E}\kern-.125emX}}

    \usepackage{xcolor}

\begin{document}

\title{Visual Pattern Recognition with on On-chip Learning: towards a Fully Neuromorphic Approach}

\author{\IEEEauthorblockN{Sandro Baumgartner, Alpha Renner,  Raphaela Kreiser, Dongchen Liang, Giacomo Indiveri,  Yulia Sandamirskaya}
  \IEEEauthorblockA{\textit{Institute of Neuroinformatics}
    \textit{University of Zurich and ETH Zurich}, Zurich, Switzerland\\
    bausandr@student.ethz.ch, alpren@ini.uzh.ch, rakrei@ini.uzh.ch, dongchen@ini.uzh.ch, giacomo@ini.uzh.ch, ysandamirskaya@ini.uzh.ch}
}

\maketitle

\begin{abstract}
We present a spiking neural network (SNN) for visual pattern recognition with on-chip learning on neuromorphic hardware. We show how this network can learn simple visual patterns composed of horizontal and vertical bars sensed by a Dynamic Vision Sensor, using a local spike-based plasticity rule. During recognition, the network classifies the pattern's identity while at the same time estimating its location and scale. We build on previous work that used learning with neuromorphic hardware in the loop and demonstrate that the proposed network can properly operate with on-chip learning, demonstrating a complete neuromorphic pattern learning and recognition setup. Our results show that the network is robust against noise on the input (no accuracy drop when adding 130\% noise) and against up to 20\% noise in the neuron parameters. 
\end{abstract}


\section{Introduction}
Convolutional neural networks (CNNs) are the state of the art approach for image recognition. Trained on suitable datasets, they enable recognition of hundreds of object classes with high precision~\cite{lecun2015deep}. However, in dynamic pattern recognition applications that use event-based cameras such as the Dynamic Vision Sensor (DVS)~\cite{Lichtsteiner2006}, the conventional CNN based approach undermines the DVS's advantages: the low latency and power consumption~\cite{Gallego2019}. As spiking neural networks (SNNs) match the event-driven nature of the DVS output, they are a natural choice to process the event-based visual output.  These networks can run efficiently in neuromorphic hardware -- computing hardware that implements SNNs on-chip~\cite{FurberEtAl2012, Chicca_etal14, Moradi2018, Davies2018}, but they cannot be easily trained with backpropagation learning rules, as in conventional CNNs.  

There are several ways of how an SNN can be configured to solve a pattern recognition task. A CNN-to-SNN conversion toolbox~\cite{rueckauer2017conversion} can be used to convert a trained CNN to an SNN that one can fine-tune for the hardware. Alternatively, several local learning methods that approximate backpropagation are being explored with spiking networks, which show promising results~\cite{Lee2016,neftci2019surrogate,zenke2018superspike}. However, these methods share the same problems of backpropagation, as they require a large amount of data for training and need retraining with the whole dataset to learn new patterns. Other approaches propose to learn a  hierarchy of feature-detectors using so-called \emph{time surfaces} to detect spatio-temporal event-patterns~\cite{Lagorce2015b,Lagorce2016}.  Unsupervised learning has also been demonstrated to show promising results in a shallow SNNs~\cite{Diehl2015, Kreiser_etal17}. 

In this line of work, \cite{Liang2019} has proposed a method of learning an SNN for pattern recognition using local learning rules -- spike-based Hebbian learning -- that are typically available in neuromorphic hardware. This work used a mixed-signal neuromorphic device DYNAP~\cite{MoradiEtAl2017} that did not support on-chip learning. The learning algorithm was run on a computer with the DYNAP chip in the loop~\cite{Liang2019}. In that work, the authors demonstrated properties of the network that go beyond standard CNNs: one-shot learning, scale- and location invariant recognition with simultaneous estimation of the scale and location of the patterns, autonomous arbitration of learning and recognition phases with detection of the novelty of the presented pattern. These properties are attractive features for neuromorphic behaving systems and for sensory-processing applications that require online learning. Here we extend that work by proposing an improved arbitration mechanism between learning and recognition, a scaling mechanism that requires fewer neurons and synapses, as well as a spike-based Hebbian learning rule that is implemented directly on a neuromorphic hardware platform without requiring a computer in the loop. We implemented the SNN pattern recognition architecture on Intel's neuromorphic research chip Loihi ~\cite{Davies2018} and replicated the results of~\cite{Liang2019} with online learning in hardware. Furthermore, we validated the robustness of pattern recognition against noise on the input and noise in neurons, as can be observed in mixed-signal neuromorphic devices. Finally, we estimated the resources needed to extend the SNN to perform a larger-scale recognition task.








\section{Hardware Setup}

To obtain visual data, we used an event-based DVS -- the DAVIS 240C~\cite{Brandli2014}. Unlike a frame-based camera, an event-based DVS does not output a frame of pixel values proportional to light intensity but emits  on- and off-events in response to local brightness changes~\cite{Lichtsteiner2006}. As the DVS responds only to changes in the visual scene, and since the DAVIS was in a fixed setup, the patterns that we presented to the camera were jiggled by hand. The generated events are transmitted to a host PC for data analysis using the address-event representation (AER)~\cite{Boahen2004} and can be displayed and processed using the jAER software toolchain\cite{jAER}. In our setup, we captured only the on-events and discarded the (redundant) off-events. The addresses of the events were downsampled on the PC from $240\times180$ to $16\times16$ pixels before sending them to the Loihi chip. In our experiments 0.25ms of the DVS recording correspond to 1 timestep on Loihi, creating input patterns, as shown in Fig.~\ref{fig:network layout}(a).

 Loihi  is a  neuromorphic research chip developed by Intel that implements SNNs on a hardware level~\cite{Davies2018}. Each of its 128 cores integrates 1,024 neural units called compartments. The compartments' behavior follows the leaky integrate and fire model. Loihi approximates the continuous-time dynamics of biological spiking neurons using a fixed-size discrete timestep model. Each compartment can be connected to any other compartment by synapses. Programmable synaptic learning rules enable online learning~\cite{Lin2018}. The downsampled visual input is provided to the network on Loihi using spike generators. These are ports connected to compartments that can emit spikes at precise timesteps. The network output spikes and weight changes are read out and sent off-chip using ``probes'', a built-in feature on Loihi to read out internal variables.

\section{Spiking Neural Network Architecture}

\begin{figure*}
  \center
  \includegraphics[width=0.95\linewidth]{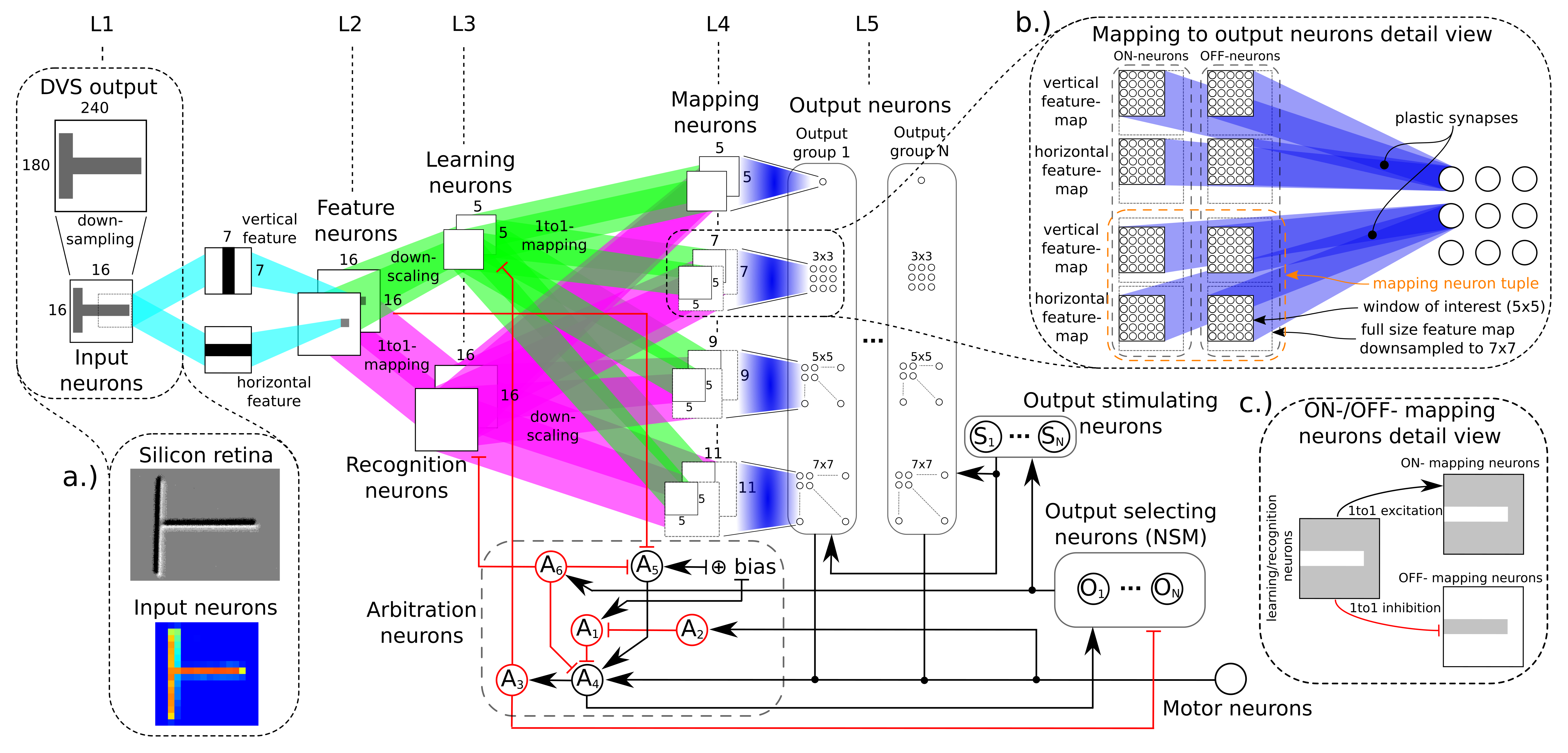}
  \caption{\textbf{The SNN architecture}. The network consists of five layers: L1 -  input layer, L2 - feature layer, L3 - learning/recognition layer, L4 - mapping layer, and L5 - output layer. The learning pathway is denoted green, the recognition pathway -- pink. Six arbitration neurons (A$_1$-A$_6$) coordinate activation of learning and recognition, a cascade of output selecting neurons (a neural state machine \cite{LiangNSM}) triggers learning of a new pattern.
 \textbf{Insets}: \textbf{(a)} Visual output of the DVS and the corresponding input neuron activity. \textbf{(b)} Each group of four (2 features, ON/OFF) 5x5 arrays of mapping neurons projects to a distinct output neuron. \textbf{(c)} ON-mapping neurons receive 1-to-1 excitation from the learning/recognition neurons; OFF-mapping neurons have a positive bias current and receive 1-to-1 inhibition from the learning/recognition neurons.}
  \label{fig:network layout}
\end{figure*}


The SNN for pattern recognition is shown in  Fig.~\ref{fig:network layout}: the input neurons (L1) project in a convolutional manner to the feature neurons (L2), using two 7x7 stride-1 kernels that detect horizontal and vertical bars. The feature neurons project to the mapping neurons (L4) via a learning and a recognition pathway (L3). Learning takes place in the plastic synapses (dark blue) between the mapping neurons and the output neurons (L5). The output neurons are assigned to output groups; per output group, one pattern can be stored. A motor neuron deactivates learning when input in front of the DVS is moving significantly. In our setup, we activate the motor neuron manually. The different parts of the SNN architecture are presented below:

\subsection{Arbitration mechanism}

The arbitration mechanism consists of six arbitration neurons (A\textsubscript{1}, ..., A\textsubscript{6}). The purpose of this neuronal circuit is to inhibit either the learning (green) or the recognition (pink) pathway. We distinguish the following five cases:

\textit{ - A new pattern is stably presented at the input:} The output neurons only spike weakly because the network has not learned the pattern before. The neuron A\textsubscript{1} has a weak positive bias input and is active if not actively inhibited. Neuron A\textsubscript{1} inhibits neuron A\textsubscript{4}. Consequently, neuron A\textsubscript{3}  also does not spike. When A\textsubscript{3} and A\textsubscript{4} are silent, one output selecting neuron (O\textsubscript{x}) gets activated and learning is triggered.

\textit{ - Learning has been triggered:} One of the output selecting neurons (O\textsubscript{x}) starts spiking. This excites A\textsubscript{6}, which  inhibits the recognition pathway as well as A\textsubscript{4}. The learning continues until O\textsubscript{x} stops spiking.

\textit{ - An already trained pattern is presented at the input:} As the network has already been trained on the presented pattern, some output neurons spike with a high rate. The output neurons  excite A\textsubscript{4} and, as a consequence, also A\textsubscript{3}. This inhibits the output selecting neurons and the learning pathway.

\textit{ - The DVS is being moved substantially, no stable input is presented:} The motor neuron spikes, activating A\textsubscript{3} and A\textsubscript{4}, which inhibit output selecting neurons and learning pathway.

\textit{ - Too few feature neurons' spikes:} A\textsubscript{5} is only weakly inhibited, and its positive bias current causes it to spike. As a consequence, A\textsubscript{3} and A\textsubscript{4} are activated, which inhibits the output selecting neurons as well as the learning pathway. This ensures that if no feature can be detected in the input, no learning is triggered.


\subsection{Spike-based learning rule}

Initially, all plastic synapses connecting the mapping neurons to the output neurons have weight 0, such that no output spikes occur before learning is triggered. Whenever learning is triggered, the output neurons of one output group are activated. For each postsynaptic spike, the synaptic weight is updated according to the following Hebbian-like learning rule:
\begin{equation}\label{eq:learning}
\Delta w(t) = \big({ x }_{ 1 }(t)-\alpha \big)\cdot \big({ w }_{ max }-w(t) \big)\cdot \big(w(t)-{ w }_{ min } \big) \cdot \lambda \quad.
\end{equation}
 Here, ${ x }_{ 1 }(t)$ is the presynaptic trace at timestep $t$ which increases with a presynaptic spike and decays exponentially over time. For ${ x }_{ 1 }(t)>\alpha$, the weight update $\Delta w(t) > 0$ and the synaptic weight grows and becomes excitatory. Synapses with ${ x }_{ 1 }(t) < \alpha$ decrease and become inhibitory. The learning stops as soon as $w(t)$ has reached ${ w }_{ max }$ or ${ w }_{ min }$ (${ w }_{ max } \in \mathbb{Z}_{>0}$, ${ w }_{ min } \in \mathbb{Z}_{<0}$). Thus, synapses with high presynaptic spiking activity become excitatory, synapses with no or low presynaptic spiking activity become inhibitory (Fig.~\ref{fig:wgt learning}). A scaling factor $\lambda$ controls the speed of learning.

\subsection{Normalization of mapping neuron activity}

To robustly distinguish a new pattern from already learned ones, we need to make sure that the same ratio of excitatory and inhibitory synapses are potentiated for each pattern. Otherwise, if there were more excitatory synapses learned for one pattern than for other patterns, this would lead to a greater excitation of its corresponding output, and it would be harder to discriminate the low and high output activity, which is key to detecting a novel pattern. To achieve this homogeneity, we need to balance the number of spiking and silent neurons in each tuple of mapping neurons at each time. The mapping neurons contain an array of ON-mapping neurons that receive excitatory input from the learning/recognition neurons, and OFF-mapping neurons, which have a positive bias current but receive inhibitory input from the learning/recognition neurons. Thus, for each spiking ON-mapping neuron, its equivalent OFF-mapping neuron is silent, and vice versa, Fig. \ref{fig:network layout}(c). Thus, the overall amount of activation going to the output neuron does not depend on the number of active pixels in a pattern. 

\subsection{Scaling mechanism}

\begin{figure}
  \center
  \includegraphics[width=0.95\linewidth]{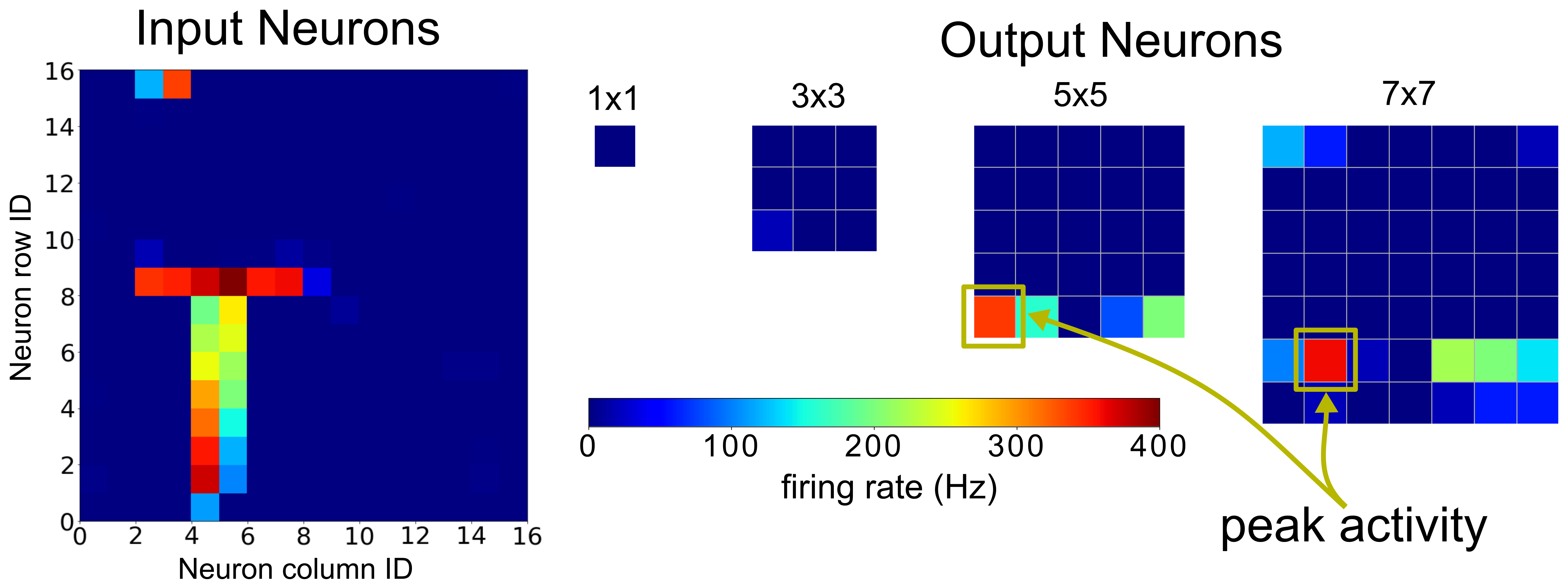}
  \caption{\textbf{Left}: The spiking activity of the input neurons when a small T-shape is presented as input. \textbf{Right}: The spiking activity of the output neurons from the output group that has learned the T-shape. The identity of the most active neuron in the output arrays represents the size and location of the input pattern.}
  \vspace{-0.5cm}
  \label{fig:output representation}
\end{figure}

To represent information about the location and size of a presented pattern on the output, within each output group, each output neuron is assigned to a certain pattern size (4 different sizes in our architecture) and location (see L5 in Fig.~\ref{fig:network layout} and Fig.~\ref{fig:output representation}). Furthermore, a distinct tuple of mapping neurons, consisting of 5x5 ON- and OFF-mapping neurons per feature, projects to each output neuron among each output group, Fig.~\ref{fig:network layout}(b). Hence, as we use 84 output neurons per output group, we also have the same amount of mapping neuron tuples, each of them corresponding to a different pattern size and location. 


During learning, a pattern is presented at full scale. In the learning pathway, the patterns are down-scaled between the feature neurons (L2) and the learning neurons (L3). For each feature, an array of 5x5 learning neurons represents the  activity of the corresponding feature neurons at a 5x5 neuron resolution.  Each of these arrays of 5x5 learning neurons projects in a one-to-one manner to the corresponding 5x5 array of ON-/OFF-mapping neurons within each mapping neuron tuple in L4. Consequently, during learning, all mapping neurons that correspond to the same feature exhibit the same  activity pattern. Thus, during learning, the same weight pattern is learned for the synapses connecting each tuple of mapping neurons to an output neuron of the active output group.


In the recognition pathway, the feature neurons (L2) project in a one-to-one manner to the recognition neurons (L3). The down-scaling takes place between the recognition neurons (L3) and the mapping neurons (L4). Each mapping neuron tuple can be assigned to a group depending on which pattern size this tuple corresponds to. As we distinguish between 4 different pattern sizes in this setup, there are four groups of mapping neuron tuples. The group corresponding to a full-size pattern contains only one mapping neuron tuple, the subsequent groups which correspond to a pattern of smaller size contain 3x3, 5x5, and 7x7 tuples respectively. For the first group of mapping neuron tuples, the output of the recognition neurons is down-scaled to 5x5 and projected to each 5x5 array of ON- and OFF-mapping neurons in a one-to-one manner. For the next three groups of mapping neuron tuples, the recognition neurons output is down-scaled to larger scales (7x7, 9x9, and 11x11) and to each tuple only a 5x5 window of interest out of this down-scaled recognition neurons output is projected. Shifting this window of interest for each tuple of mapping neurons results in size and location invariant recognition (Fig.~\ref{fig:network layout}(b)).

\subsection{Winner-take-all neural state machine}

Following \cite{Liang2019}, we use a cascade of neural state machines (NSMs)  to stimulate the output selecting neurons and trigger learning of a new pattern. The NSM network is described in \cite{Liang2018} and \cite{LiangNSM}. Whenever A\textsubscript{3} and A\textsubscript{4} are silent, a competition among  NSMs starts, which results in a winner NSM being active and pushing all other NSMs to the inactive state. The active NSMs output selecting neuron O\textsubscript{x} spikes, which stimulates its corresponding output stimulating neuron S\textsubscript{x} as well as A\textsubscript{6}. Consequently, after a short delay that gives the network time to silence the recognition pathway and to activate the learning pathway, S\textsubscript{x} starts spiking. This excites the output neurons of the corresponding output group which update their weights according to Eq.~\eqref{eq:learning}.

After the winner NSM has been in an active state for a certain amount of time, the presynaptic trace ${ x }_{ 1 }(t)$ of a plastic synapse in the NSM reaches a threshold $\alpha$ causing the synaptic weight to decrease based on the following learning rule:
\begin{equation}\label{eq:nsm learning}
\Delta w(t) = \big(\alpha-{ x }_{ 1 }(t) \big)\cdot \big({ w }_{ max }-w(t) \big)\cdot \lambda-{ x }_{ 1 }(t)\cdot \gamma.
\end{equation}
The winner NSM is inactivated and terminates the learning process. Due to the decreased weight, this NSM will not  win again. For the next new pattern, another NSM will be selected and another output group will be stimulated.

\section{Results}

\begin{figure}
  \center
  \includegraphics[width=0.85\linewidth]{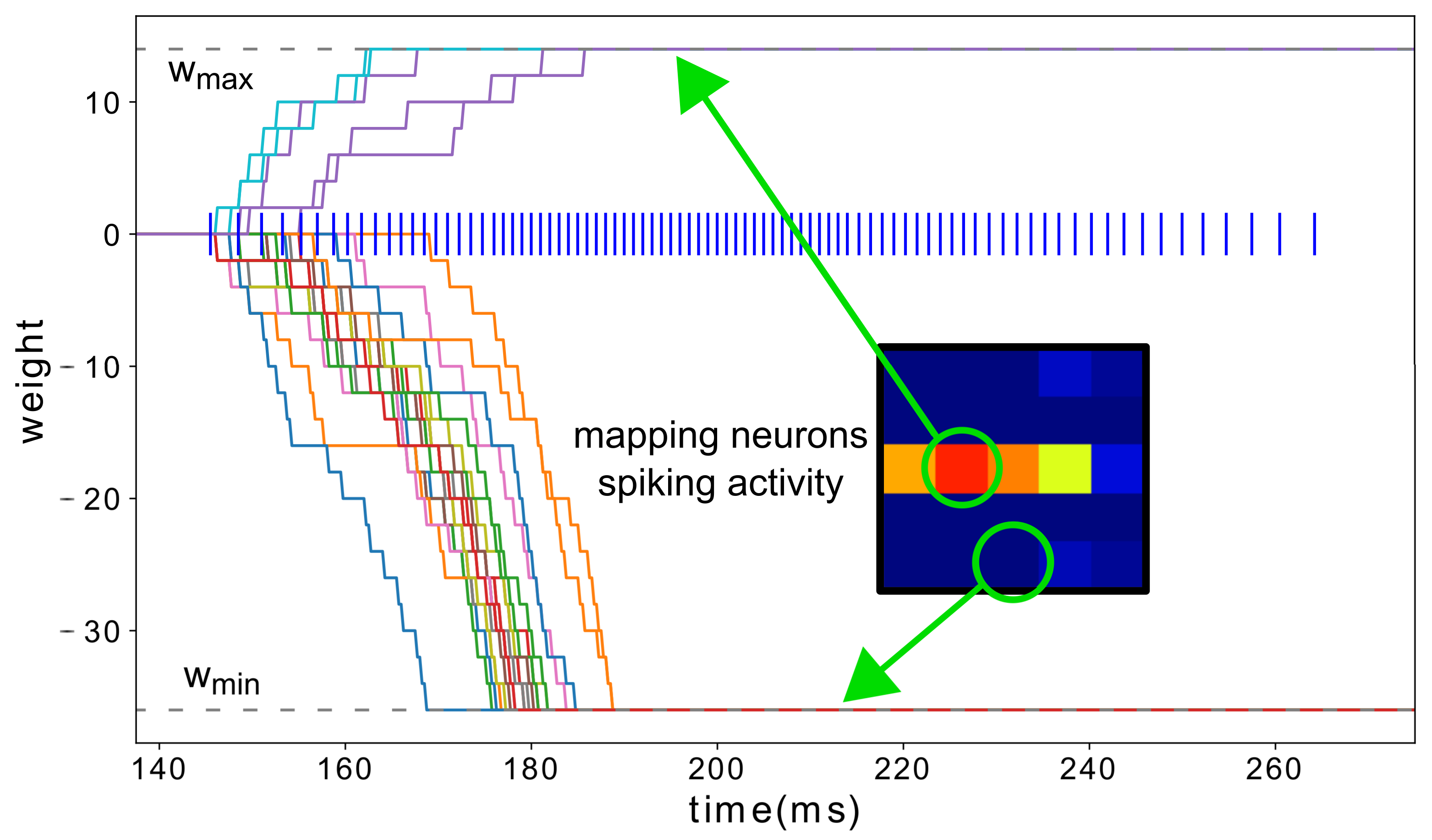}
  \caption{Weight updates in the plastic synapses connecting one mapping neuron group to an output neuron,  when a new pattern is presented. Each colored line shows the weight of a synapse over time. When the weight reaches w\textsubscript{max} or w\textsubscript{min} (grey dashed lines), the learning stops for that synapse. Blue dashes represent spikes of the output neuron.}
  \vspace{-0.5cm}
  \label{fig:wgt learning}
\end{figure}

\begin{figure}
  \center
  \includegraphics[width=\linewidth]{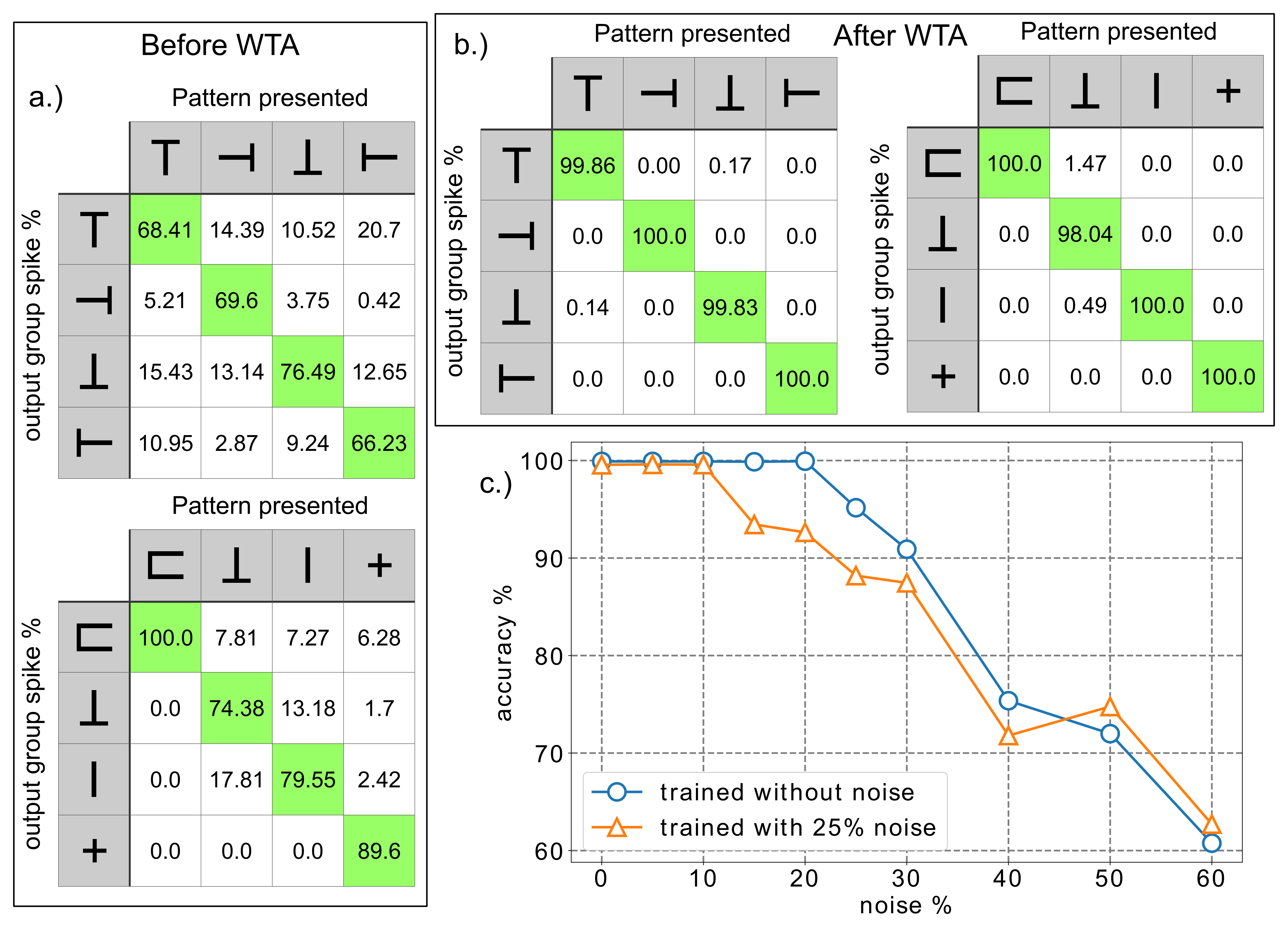}
  \caption{Confusion matrices (a) before and (b) after the WTA layer. (c) Accuracy as a function of noise injected to the feature neurons.}
    \vspace{-0.5cm}
  \label{fig:accuracy fig}
\end{figure}

We performed experiments with a network that can distinguish 4 different patterns. For training, each pattern has been presented for 1 second. For evaluation, we presented the patterns for 2.5 seconds and monitored the output spikes. The following properties of the network have been examined:

\textit{Accuracy:} For accuracy evaluation, a winner-take-all (WTA) mechanism was appended to the output of the network (not shown in Fig.~\ref{fig:network layout}). Per output group, there is a single WTA-neuron to which all output neurons of this output group project. As a result of the competition among WTA-neurons, only neuron with the strongest input from its output neurons persists spiking. The accuracy was then measured by counting the number of spikes from the output neurons and the WTA-neurons. As can be seen in Fig.~\ref{fig:accuracy fig}, the classification accuracy for 4 patterns is close to 100\% after the WTA network.

\textit{Robustness:} Even adding 130\% noise to the input layer didn't reduce the classification accuracy, which shows that the network is robust against noise on the input. To evaluate the robustness against noise in neuronal elements as it may appear in mixed-signal neuromorphic hardware, noise was injected in the feature layer. Fig.~\ref{fig:accuracy fig} shows the classification accuracy as a function of the injected feature neuron noise.

\textit{Latency:} Fig.~\ref{fig:wgt learning} shows that when a new, not previously learned pattern was presented, all plastic weights have either converged to ${w}_{max}$ or ${w}_{min}$ within 200\,ms. Thus, the one-shot learning process is completed in less than 200\,ms. To investigate how fast the output adapts to a previously trained pattern presented, we measured the time between the first time at which the new pattern is presented and the time when the output neurons spiking activity indicates that this pattern is being presented. The measurements have shown that 15-20\,ms after the presentation of the next pattern, the spiking activity of the output neurons has already adapted to the new input.

\textit{Scalability:} The network can be extended to distinguish more than four patterns by adding an additional output neuron group and an NSM per additional pattern. The number of additional neurons and synapses scales linearly with the number of patterns: 92 neurons and 8.5K synapses per pattern at the resolution and number of features used here.

\section{Conclusion and Outlook}

We proposed an SNN architecture that enables online, one-shot unsupervised learning on neuromorphic hardware. The pattern learning and recognition are robust against noisy inputs. We validated the model on a small DVS-datasets and showed promising results regarding accuracy and latency. Generalization to more patterns could be achieved on the same hardware. Rather than operating directly on the DVS output, the same network could be used to process features produced by a pre-trained CNN. Our online unsupervised learning approach could then be used to build a hardware classifier that estimates more general patterns, with a tuning to their size and location.

\section{Acknowledgements}

Funding was provided by the SNSF Project Ambizione (grant PZOOP2\_168183) and EU ERC grant NeuroAgents (Grant No. 724295). We thank Intel Labs for their support with the neuromorphic chip.

\bibliographystyle{IEEEtran}
\bibliography{references,literature_mendeley}

\end{document}